\documentclass[12pt,a4paper]{article}
\usepackage{amsfonts}
\usepackage{amsmath,amsfonts,amsthm,mathrsfs,grffile}
\usepackage{epstopdf}
\usepackage{graphicx}
\usepackage{verbatim}
\usepackage[ruled]{algorithm2e}   
\usepackage{color}
\usepackage{picinpar}
\usepackage{booktabs}
\usepackage[doublespacing]{setspace}
\usepackage[authoryear]{natbib}
\usepackage{enumerate}

\newtheorem{thm}{Theorem}

\newtheorem{lem}{Lemma}

\theoremstyle{definition}   %

\theoremstyle{remark}

\numberwithin{equation}{section}

\newcommand{\be}{\mathbf{e}}

\newcommand{\bX}{\mathbf X}
\newcommand{\bh}{\mathbf{h}}
\newcommand{\bw}{\mathbf{w}}

\newcommand{\bs}{\mathbf{s}}
\newcommand{\bx}{\mathbf{x}}
\newcommand{\bLambda}{\boldsymbol{\Lambda}}

\newcommand{\bepsilon}{\boldsymbol{\epsilon}}
\newcommand{\bA}{\mathbf{A}}
\newcommand{\bL}{\mathbf{L}}
\newcommand{\bD}{\mathbf{D}}
\newcommand{\bU}{\mathbf{U}}
\newcommand{\bH}{\mathbf{H}}
\newcommand{\bW}{\mathbf{W}}

\def\eop{{\hfill\vbox{\hrule height .3pt
      \hbox{\vrule width.3pt height 7pt
      \kern 7pt
      \vrule width .3pt}
      \hrule height .3pt}} \par\bigskip}

\title{Generalization bounds for graph convolutional neural networks via Rademacher complexity }

\author{Shaogao Lv$^{\dag}$
\\
{\small $^{\dag}$Center of Statistics and Data Science, Nanjing Audit University, Nanjing, China;} \\
}
\date{}

\begin{document}
\maketitle

\setcounter{page}{1}
\begin{abstract}
This paper aims at studying the sample complexity of graph convolutional networks (GCNs), by providing  tight
 upper bounds of Rademacher complexity for GCN models with a single hidden layer.  Under regularity conditions,  theses derived complexity bounds explicitly depend on the largest eigenvalue of graph convolution filter and the degree distribution of the graph. Again, we provide a lower bound of Rademacher complexity for GCNs to show optimality of our derived upper bounds.  
Taking two commonly used examples as representatives,  we discuss the implications of our results in designing graph convolution filters an graph distribution.
\end{abstract}


\section{Introduction}
Graph neural networks (GNNs) \citep*{Scarselli2009,Bronstein2017,Hamilton2017} have become one of the most important branches in various deep learning, due to their remarkable power in learning with graph-structured data. Specially, GNNs have achieved excellent empirical performances in various real-world applications, such as social networks \citep*{Hamilton2017}, physical systems \citep*{Battaglia2016}  and biological analysis \citep*{Fout2017} and knowledge graphs \citep*{Ying2018}.
The research of graph neural networks can be traced back to the work in \citep*{Sperduti1997}, having gained rapid developments and extensions recently inspired from spectral graph \citep*{Laffey2008} and graph convolution in \citep*{Bruna2014}.
A thorough review of different graph neural network models and their applications was presented in \cite{Zhou2019}.

Despite the numerical success, the theoretical understanding of the underlying learning mechanisms of GNNs is still limited in the literature \citep*{Verma2019,Du2019,Garg2020}. 
In this paper, we are primarily concerned with generalization and consistency of GCNs with a single hidden layer, given that GCNs are a family of basic and commonly-used graph models  and inherent some advantages (e.g. local representation and shared parameters ) from the well-known convolutional neural networks on grid-type datasets \citep*{Bruna2014}.  
Among several topics of  learning theory, characterizing generalization behavior of a learning system
 is one of fundamental issues in the machine learning literature. One of the major challenges involving GCNs is 
 their generalization bounds can largely depend on some specific configurations on graphical structures (e.g., spectra of graphs), as well as various regularizations appeared in Euclidean-spaces related learning problems \citep*{Jakubovitz2019}.  
 
In classical learning theory, the capability of a function space contained in the learning system is closely related to the generalization behavior for its learning rule, and particularly  Rademacher complexity \citep*{Bartlett2001} is a standard tool to evaluate the generalization bound on the function class. Precisely,  the Rademacher complexity can be used to obtain uniform convergence results over a hypothesis set of possible rules.

Over the last several years, there is a long line of learning theory works on the generalization and consistency results of
conventional deep neural networks \citep*{Bartlett2007,Neyshabur2015,Golowich2019,Jakubovitz2019}. More specifically, they show that generalization can be guaranteed theoretically as long as the norms of those learned parameters are bounded and  deep
neural networks satisfy certain constraints on network width and depth.
From an algorithm-based perspective, a network predictor  also provably generalizes well if only it follows a stable or robust property in some senses with respect to data perturbation \citep*{Hardt2016}.  However, most of those results focus on conventional neural networks (e.g. feedforward neural network and convolutional neural network) on regular data. 
For generalization analysis of GNNs, we expect that its  generalization bound can leverage  the role of graph structures  in theory. 

In contrast to the aforementioned standard neural networks over Euclidean spaces, only few works explore the reason why GNNs generalize  well to unseen graph-structured data. For example, \citet*{Verma2019} first explored generalization bounds of stochastic gradient for GCNs via the notation of algorithmic stability. \cite{Zhang2020} derived fast convergent rates of an accelerated gradient descent for a single-layer GNN when the ground truth has zero-generalization error. However, the two work mentioned above are restricted to a specific model variant with no hidden layers, resulting in the lack of rich feature representation.
Moreover, they only focused on specific optimization algorithms such as stochastic gradient descent and its variant. 

 We also notice that, 
 reference \cite{Du2019} analyzed the generalization ability of GNNs through graph neural tangent kernel models, which cannot identify the superiority of neural network over kernel methods.  
 Very recently, only for graph-focused task where each sample is a triplet of node features, edge matrix and output label,  \cite{Liao2020} employed a PAC-Bayesian approach to derive generalization bounds for deep GCNs and message passing GNNs. This improves
 the existing result established in \cite{Garg2020} under similar settings. Yet, the primary difference of our work from \cite{Liao2020} and \cite{Garg2020} is that we  consider the specific node-focused task of GCNs, which only involves a fixed adjacency matrix. Unlike the graph-focused task,   the data over the node-focused task is no longer subject to the standard i.i.d assumption.

{\bf Contributions.} This paper derives generalization bounds of a GCN model with one hidden-layer in a semi-supervised learning problems.  The generalization performance is guaranteed via providing sharp upper bounds of Rademacher complexity of the hypothesis set.  In particular,  we show that the generalization bound of the GCN model depends upon the largest absolute eigenvalue of its graph convolution filter, as well as the degree distribution of the graph. Furthermore, we provide a lower bound of the Rademacher complexity of the GCN model, which thereby shows the optimality of our derived upper bounds.

The rest of the paper is organized as follows. Section 2 introduces the problem formulation. Section 3 presents the main theoretical results. The detailed proofs are provided in Section 4. In Section 5,  we list two classes of  popular graph convolution filters to show the implications of our consistency for GCNs. The paper is concluded in Section 6 with a short discussion of future work.

\section{Problem Formulation}
Learning with graph structures attempts at designing an efficient algorithm with integrating attribution features and structural features in a unified manner.
Various graph methods have been proposed  in the literature for fully exploring graph-structured data. The literature includes many typical examples, such as Deepwalk \citep*{Perozzi2014}, graph convolutional neural network \citep*{Bruna2014} and  message passing based GNNs \citep*{Li2015}. This paper focus on Rademacher complexity and generalization bounds for GCNs with a hidden-layer. 

In this paper, an undirected graph is represented as $\mathcal{G}=\{\mathcal{V},\mathcal{E}\}$, where $\mathcal{V}=\{\nu_1,\nu_2,...,\nu_n\}$ is the
set of nodes with size $|\mathcal{V}|=n$ and $\mathcal{E}\subset \mathcal{V}\times \mathcal{V}$ is the set of $|\mathcal{E}|$ edges.  Let $\bA=(a_{ij})\in \mathbb{R}^{n\times n}$ be the adjacency matrix of $\mathcal{G}$ with added self-connections, such that $a_{ij}=1$ if and only if there exists an edge between node $\nu_i$ and node
$\nu_j$, $i, j \in [n]$, and $a_{i,i}=1$ for all $i \in [n]$. 
For GCNs, we are interested in  node-focused tasks for the standard semi-supervised learning problem, where all sample points are assigned over all nodes of a fixed graph. Under the semi-supervised setting, we denote by $\Omega \subset \mathcal{V}$  the set of node indices with observed labels such that $m:=|\Omega|< n$, and let $\Omega^c$ be its complementary set.
Precisely, each node $\nu_i$ in $\Omega$ corresponds to an input feature vector, denoted by $\bx_i \in \mathcal{X}\subset \mathbb{R}^d$  and a label $y_i \in \mathcal{Y}\subset \mathbb{R}$.  The objective of the GNN is to predict $y_i$ for every $i\in \Omega^c$.

For an undirected graph, its Laplacian matrix, $\bL\in \mathbb{R}^{n\times n}$ is defined as $\bL=\bD-\bA$, where
$\bD\in \mathbb{R}^{n\times n}$ is a degree diagonal matrix whose diagonal element $d_{ii}=\sum_{j}a_{ij}$ for all $i\in[n]$.
Since $\bL$ is symmetric,  the eigen-decomposition of $\bL$ is endowed with the form: $\bL=\bU\bLambda\bU^T$ with $\bLambda=\hbox{diag}(\lambda_1,\lambda_2,...,\lambda_n)$ consisting of the Laplacian eigenvalues. 

A graph filter $g(\bL)$ is defined as a function on the graph Laplacian $\bL$, playing a crucial role in efficiently representing  graph-structured data. Given a vector $\bs\in \mathbb{R}^n$, where $s_i\in \mathbb{R}$ is the scalar feature of node
$i$, the filter operation of $\bs$ with a graph convolution filter $f(\bL)$ is defined as:
\begin{align}\label{conv}
\tilde{\bs}=g(\bL)\bs=\bU g(\bLambda)\bU^T\bs,
\end{align}
where we use the conclusion that $g(\bL)=g(\bU\bLambda\bU^T)=\bU g(\bLambda)\bU^T$ according to  orthogonality of $\bU$. 
We assume that  $g(\bL)$ belongs to a graph shift operator, satisfying $[g(\bL)]_{ij}=0$ if $i\neq j$ or $e_{ij}\notin \mathcal{E}$.
In other words, the entry of $g(\bL)$ can be nonzero only for the diagonal place or all the edge coordinates. 
Essentially, the graph convolution of a given vector is equivalent to weight-averaging the feature of each node with its neighbors.

A multi-layer GNN uses the following classic iterative scheme for updating the feature representation:
\begin{align}\label{multrep}
\bH^{(t+1)}=\sigma\big(g(\bL)\bH^{(t)}\bW^{(t+1)}\big)
\end{align}
where $\bH^{(t+1)}\in \mathbb{R}^{n\times m_{t+1}}$ is the node feature representation output by the $(t+1)$-th GCN layer,
$\bW^{(t+1)}\in \mathbb{R}^{m_t\times m_{t+1}}$ represents the weight matrix of the $(t+1)$-the GCN later, and $\sigma(\cdot)$
is a component-wise nonlinear activation function. Specially, the initial node representation are $\bH^0=\bX$ and $m_0=d$.
In this paper, we focus on GCN with a single hidden layer and an output layer with a single neuron. In this specific setup, 
Equation \eqref{multrep} can be written as:
\begin{align*}
\bh^{(2)}=\sigma\Big(g(\bL)\sigma\big(g(\bL)\bX\bW^{(1)}\big)\bw^{(2)}\Big),
\end{align*}
where $\bh^{(2)}\in \mathbb{R}^n$ and $\bw^{(2)}\in \mathbb{R}^{k}$. For the node-focused tasks,  the output $f_i$ of the node $\nu_i$ of the GCN can be rewritten as
\begin{align}\label{functionform}
f_i:=h_i^{(2)}=\sigma\left(\sum_{t=1}^kw_t^{(2)}\sum_{v=1}^n[g(\bL)]_{iv}\times 
\sigma\Big(\sum_{l=1}^dw_{lt}^{(1)}\sum_{j=1}^n[g(\bL)]_{vj}x_{jl}\Big)\right).
\end{align}

Theoretically, the risk of $f$ over the population probability is measured by 
$$
\mathcal{E}(f):=\mathbb{E}_{\mathcal{X}\times \mathcal{Y}}[\ell(y,f(\bx))],
$$
where $\ell$ is the loss function defined as a map:
$\ell: \mathcal{Y}\times \mathcal{Y}\rightarrow \mathbb{R}^+$. The best predictor $f^*$ is a global minimizer of $R(f)$ over a  hypothesis space $\mathcal{F}$, denoted by
$f^*=\arg \min_{f\in \mathcal{F}} R(f)$.
 
Given a  training set $Z^m=\{(\bx_1,y_1),(\bx_2,y_2),...,(\bx_m,y_m)\}$ over $\Omega$ and the graph filter $g(\bL)$, the learning objective instead is to estimate parameters $(\bW^{(1)},\bw^{(2)})$ based on $Z^{m}$ and $g(\bL)$. Concretely, we attempts to  minimize the empirical risk functional over $\mathcal{F}$
$$
\mathcal{E}_m(f):=\frac{1}{m}\sum_{j=1}^m\ell(y_j,f(\bx_j)).
$$
Note that the empirical risk term can be induced by partial likelihood estimation given the input points in statistics.
Recall that, 
a predictor $h$ is said to generalized if  for any $\varepsilon>0$, the following holds with probability approaching 
one as $m\rightarrow \infty$
$$
\mathbb{P}\Big(\mathcal{E}(f)\geq \mathcal{E}_m(f)+\varepsilon\Big)\rightarrow 0,
$$
where the probability is over the randomness of $Z^m$.

A predictor with generalization guarantee is closely related to the complexity of its hypothesis space.  We adopt the Rademacher complexity to measure the functional complexity. For a function set $\mathcal{F}$ defined over the graph $\mathcal{G}$,  the  empirical Rademacher complexity  is defined as
$$
\widehat{\mathcal{R}}(\mathcal{F}):=\mathbb{E}_{\epsilon}\Big[\frac{1}{m}\sup_{f\in \mathcal{F}}\Big|\sum_{j=1}^m\epsilon_j f(\bx_j)\Big|\bx_1,\bx_2,...,\bx_n\Big]
$$
where $\{\epsilon_i\}_{i=1}^m$ is an i.i.d. family (independent of $(\bx_i)$) of Rademacher variables. Note that the conditional expectation here is taken with respect to $\{\epsilon_i\}_{i=1}^m$ given that $\{\bx_i\}_{i=1}^n$ is fixed, not limited to the supervised input data over $\Omega$.

Since the neighbor representation of graph shift operators is maintained, the output of the first layer in Equation \eqref{functionform} can be written as a vector form
\begin{align}\label{fisrtequ}
\sigma\Big(\sum_{l=1}^dw_{lt}^{(1)}\sum_{j=1}^n[g(\bL)]_{vj}x_{jl}\Big)=\sigma\Big(\sum_{j\in N(v)}[g(\bL)]_{vj} 
\big\langle\bx_j, \bw_t^{(1)} \big \rangle \Big),
\end{align}
where $N(v)$ denotes the set of neighbors of $v$, and 
 $\bW=(\bw_1,\bw_2,...\bw_k)$ represented in a column-wise manner. Thus, the class of functions defined over the  node set $\Omega$  with
norm constraints  coincides with
\begin{align}\label{hypothesis}
\mathcal{F}_{D,R}:=\Big\{f(\bx_i)&=\sigma\Big(\sum_{t=1}^kw_t^{(2)}\sum_{v=1}^n[g(\bL)]_{iv}\times \sigma\big(\sum_{j\in N(v)}[g(\bL)]_{vj} 
\big\langle\bx_j, \bw_t^{(1)} \big\rangle \big) \Big),\,i\in[m],\nonumber\\
&\hspace*{0.7cm}\|\bW^{(1)}\|_{F}\leq R,\,\|\bw^{(2)}\|_2\leq D \Big\},
\end{align}
where the Frobenius norm  of a matrix  is given as $\|\bW\|_{F}^2:=\sum_{ij}W_{ij}^2=\sum_{t=1}^k\|\bw_t\|_2^2$.  Bounding the population Rademacher complexity over  $\mathcal{F}_{D,R}$ is quite challenging, mainly due to the fact that
each output $h_i^{(2)}$ depends on all the input features that are connected to node $\nu_i$, as shown in the functions in \eqref{functionform} and \eqref{hypothesis}.


\section{Main Results}
In this section, we present the main results: upper bounds and lower bounds of Rademacher complexity of GCNs with one-hidden layer.
We then relate them to the generalization bounds of GCNs using some existing lemma on uniform convergence.

In this paper,  consider the class $\mathcal{F}_{D,R}$ of two-layer graph neural networks over $\mathbb{R}^d$, and particularly
the following technical assumptions are required in theory. 
\begin{enumerate}[(1)]
	\item The activation function $\sigma(\cdot)$ is $L$-Lipschitz continuous with some $L>0$, and also satisfies $\sigma(0)=0$ and $\sigma(\alpha z)=\alpha \sigma(z)$ for any $\alpha\geq 0$.
	\item  The $L_2$-norm of the  feature vector in the input space is bounded, namely, for some constant $B>0$, we assume  $\|\bx_i\|_2\leq B$ for all $i$.	
	\item The Frobenius norm of the weight matrix in the first layer of the GCN class is also bounded, namely, $\|\bW^{(1)}\|_F\leq R$ with some constant $R>0$.
	\item The $L_2$-norm of the weight vector of the output is bounded as well, namely, $\|\bw^{(2)}\|_2\leq D$ with some constant $D>0$.
	\item The number of neighbors of each node is equal to each other, namely, for some common constant $q\in \mathbb{N}^+$, assume $q:=N(i)$ for all node $\nu_i\in \Omega$.
\end{enumerate}

The above assumptions are common in the neural network and graph literature. In particular, Condition $(1)$
indicates that the activation function is positive-homogeneous, including the popular ReLU as special case.
Conditions $(2)-(4)$ on norm constraints of parameters and input data allow $\mathcal{F}_{D,R}$ to grow within a compact metric space. Finally, the last condition requires us to focus on homogeneous graphs, such as Erdos-R$\acute{e}$nyi graphs and regular graphs. 

Let $\widetilde{\bX}_v=(\tilde{\bx}_1^T,...,\tilde{\bx}_q^T)^T \in \mathbb{R}^{q\times d}$ be the feature matrix of the nodes in $\mathcal{G}_v$, where all $\tilde{\bx}_i$'s are denoted to be  reordered input data according the neighbors of node $\nu$.
We are now equipped to state our main results.

\begin{thm}\label{upperres}
	Let $\mathcal{F}_{D,R}$ be a class of real-valued GCNs with one hidden layer over the domain $\mathcal{X}\in \mathbb{R}^d$, where all the components in $\mathcal{F}_{D,R}$ satisfy these assumptions given as above.  Then
	$$
	\widehat{\mathcal{R}}(\mathcal{F}_{D,R})
	\leq 8L^2BDR|\max_{k\in [q]}\big\{ \big\|\widetilde{\bX}_q[g(\bL)]_{\cdot k} \big\}\sqrt{\frac{1}{m}}\sum_{l=1}^q  \max_{j\in[m]}\big|[g(\bL)]_{jn_l(j)}\big|,
	$$
	In addition, we assume that a fixed graph filter $g(\bL)$ has a finite maximum absolute eigenvalue,  denoted by $\lambda_{\hbox{max}}(\mathcal{G})$. Then 
	$$
	\widehat{\mathcal{R}}(\mathcal{F}_{D,R})
	\leq 8L^2BDR|\lambda_{\hbox{max}}(\mathcal{G})|\sqrt{\frac{q}{m}}\sum_{l=1}^q  \max_{j\in[m]}\big|[g(\bL)]_{jn_l(j)}\big|,
	$$
	where $n_l(j)$ refers to the $l$-th neighbor of  node $\nu_j$ by a given order. 
\end{thm}
Theorem \ref{upperres} indicates that the upper bound of $	\widehat{\mathcal{R}}(\mathcal{F}_{D,R})$ depends upon the number of label, the degree distribution of the graph, and the choice of the graph convolution filter. It is interesting to observe that, the above bound is independent of  the graph size $(n)$ for the traditional regular graphs, which will be discussed in details in Section 5. 

It is also worth noting that, for the two-layer neural network with the width $k$, our upper bound only has an explicit dependence of the Frobenius norm of parameter matrix, while is independent of the network width.  We hope that such a conclusion can be extended to deep GCNs in future work.

\begin{thm}\label{lower}
	Let $\mathcal{F}_{D,R}$ be a class of GCNs with one-hidden layer, where the parameter matrix and the parameter vector satisfy 
	$\|\bW^{(1)}\|_F\leq R$  and $\|\bw^{(2)}\|_2\leq D$ respectively. Then there exist a choice of $L$-Lipschitz activation function, data points $\{\bx_{i}\}_{i=1}^n$ and a family of given graph convolutional filters, such that
	$$
	\widehat{\mathcal{R}}(\mathcal{F}_{D,R})\geq
	\frac{L^2BDR}{\sqrt{m}}\min_{k\in [q]}\Big\{ \big\|\widetilde{\bX}_q[g(\bL)]_{\cdot k} \big\|_2\sum_{t=1}^q[g(\bL)]_{kt}\Big\}.
	$$
\end{thm}
We would like to compare the above lower bound with the upper bound of Theorem \ref{upperres}. In fact, if the minimax-term in Theorem \ref{lower} matches the maximum-term in Theorem \ref{upperres}, this shows that our derived upper bound is tight up to some constants.

As an application of our results in Rademacher complexity of the GCN models to generalization analysis, 
we now state the fundamental result on the generalization bound involving Rademacher complexity \citep*{Bartlett2001}.

\begin{lem}\label{radegener}
	Assume that the input feature is fixed. For any $\delta>0$, with probability at least $1-\delta$, 
	$$
	\mathcal{E}(h)\leq \mathcal{E}_n(h)+2\widehat{\mathcal{R}}(\mathcal{H})+\sqrt{\frac{2\log(2/\delta)}{n}},\quad \forall\,h\in \mathcal{H},
	$$ 
	and also, with probability at least $1-\delta$, 
$$
\mathcal{E}(h)\leq \mathcal{E}_n(h)+2\mathbb{E}_{\mathcal{Z}}\big[\widehat{\mathcal{R}}(\mathcal{H})\big]+\sqrt{\frac{\log(1/\delta)}{2n}},\quad \forall\,h\in \mathcal{H}.
$$ 	
\end{lem}
Remark that,  the first part of Lemma \ref{radegener}  tells us that the expected risk of any predictor  depends solely on the empirical data. We denote $\mathcal{H}=\{\ell(y,f(\cdot)), f\in \mathcal{F}_{D,R}\}$ throughput the paper. To connect  $\widehat{\mathcal{R}}(\mathcal{H})$ with 	$\widehat{\mathcal{R}}(\mathcal{F}_{D,R})$ using Lemma \ref{radegener},  we still need to the Lipschitz condition of the loss function stated as follows.  

{\bf Lipschitz-Assumption on loss function.} Assume that the loss function $\ell(y,\cdot)$ is Lipschitz-continuous,
$$
|\ell(y,f(\cdot))-\ell(y,f'(\cdot))|\leq \alpha_\ell |f(\cdot))-f'(\cdot))|,\quad \forall\,y\in \mathcal{Y}.
$$ 

By the contraction property of (empirical) Rademacher complexity \citep*{Bartlett2001}, it is known that Rademacher complexity   
$$
\widehat{\mathcal{R}}(\mathcal{H})\leq 2\alpha_\ell \widehat{\mathcal{R}}(\mathcal{F}_{D,R}).
$$
Based on the results given in Theorem \ref{upperres} and Lemma \ref{radegener}, we obtain the following generalization bound which holds for a class of homogeneous GCNs with one-hidden layer.

\begin{thm}
	Let $f: \mathcal{X}\rightarrow \mathbb{R}$ be any given predictor of a class of GCNs with one-hidden layer. Assume that Assumptions
	(1)-(5) hold and additionally the loss is Lipschitz continuous. Then for any $\delta>0$, with probability at least $1-\delta$, for any predictor $f\in \mathcal{F}_{D,R}$, we have
	$$
	\mathcal{E}(f)\leq \mathcal{E}_n(f)+16L^2BDR\alpha_\ell|\lambda_{\hbox{max}}(\mathcal{G})|\sqrt{\frac{q}{m}}\sum_{l=1}^q  \max_{j\in[m]}\big|[g(\bL)]_{jn_l(j)}\big|+\sqrt{\frac{2\log(2/\delta)}{n}}.
	$$ 
\end{thm}
Note that the Lipschitz condition of the loss is fulfilled for the hinge loss and the least square loss on a bounded domain.

\section{Proofs}
The road-map of the proof mainly consists of the following three steps.
First, we show that  upper bound of the  empirical Rademacher complexity with respect to two parameters $(\bW^{(1)},\bw^{(2)})$
can be reduced to the Rademacher bound under the $L_2$-norm only involving the parameter $\bW^{(1)}$;
Second, the Rademacher bound under the $L_2$-norm further can be converted into a standard one with removing the $L_2$-norm;
Last, our desired bound for Rademacher complexity is derived from  the graph spectral theory.

\begin{proof}
We begin with rewriting Equations \eqref{functionform} for brevity. 	Define $\bh_{v}:=[h_{1}^v,h_{2}^v,...,h_{k}^v]$, where
each $h_{t}^v\in \mathbb{R}$  is defined as 
\begin{align*}
h_{t}^v:= h^v(\bw_t^{(1)})=\sigma\Big(\sum_{j\in N(v)}[g(\bL)]_{vj} 
\big\langle\bx_j, \bw_t^{(1)} \big\rangle \Big).
\end{align*}	
Following the definition of $\bh_{v}$ and \eqref{fisrtequ}, Equations \eqref{functionform} can be rewritten as
\begin{align*}
h_i^{(2)}=\sigma\Big(\sum_{v\in N(i)}[g(\bL)]_{iv} 
\big\langle\bh_v, \bw^{(2)} \big \rangle \Big).
\end{align*}
Since the activation function $\sigma$ is $L$-Lipschitz continuous, using the contraction property of Rademacher average,
we have
\begin{align}\label{radfirst}
\widehat{\mathcal{R}}(\mathcal{F}_{D,R})&=\mathbb{E}_{\epsilon}\Big[\frac{1}{m}\sup_{f\in \mathcal{F}_{D,R}}\Big|\sum_{j=1}^m\epsilon_j f(\bx_j)\Big|\bX\Big]\nonumber\\
&\leq 2L \mathbb{E}_{\epsilon}\Big[\frac{1}{m}
\mathop{\rm{sup}}_{
	\|\bW^{(1)}\|_{F}\leq R
	\atop
	\|\bw^{(2)}\|_2\leq D}
\Big|\sum_{j=1}^m\epsilon_j \times  \sum_{v\in N(j)}[g(\bL)]_{jv} 
\big\langle\bh_v, \bw^{(2)} \big \rangle \Big].
\end{align}
Among the proof, we omit the symbol $\bX$ as in  \eqref{radfirst} for notional simplicity.
Observe that we can bound the term in the supremum in \eqref{radfirst} as follows:
\begin{align}\label{supbound}
\Big|\sum_{j=1}^m\epsilon_j \times  \sum_{v\in N(j)}[g(\bL)]_{jv} 
\big\langle\bh_v, \bw^{(2)} \big \rangle \Big|&=
\Big| \Big\langle\sum_{j=1}^m\epsilon_j   \sum_{v\in N(j)}[g(\bL)]_{jv} \bh_v, \bw^{(2)} \Big \rangle \Big|\nonumber\\
&\leq D \times \Big\|\sum_{j=1}^m\epsilon_j   \sum_{v\in N(j)}[g(\bL)]_{jv} \bh_v \Big\|_2,
\end{align}
where the inequality follows from Cauchy-Schwartz inequality and  $\|\bw^{(2)} \|_2\leq D$. 
Substitute \eqref{supbound} into \eqref{radfirst}, one gets
 \begin{align}\label{radsecond}
 \widehat{\mathcal{R}}(\mathcal{F}_{D,R})
 \leq \frac{2LD}{m}\mathbb{E}_{\epsilon}\Big[\mathop{\rm{sup}}_{\|\bW^{(1)}\|_{F}\leq R}
 \Big\|\sum_{j=1}^m\epsilon_j   \sum_{v\in N(j)}[g(\bL)]_{jv} \bh_v \Big\|_2\Big].
 \end{align}
The above inequality shows that the complexity bound for depth $2$ networks is reduced to a complexity bound for depth $1$ networks under the $L_2$-norm. Note that the activation function is positive-homogeneous by assumption, that is, $\sigma(\alpha z)=\alpha \sigma(z)$ for any $\alpha \geq 0$ and $z\in \mathbb{R}$. Then
\begin{align*}
\Big\|\sum_{j=1}^m\epsilon_j   \sum_{v\in N(j)}[g(\bL)]_{jv} \bh_v \Big\|_2^2&=\sum_{t=1}^k\Big(\sum_{j=1}^m\epsilon_j
 \sum_{v\in N(j)}[g(\bL)]_{jv}h_{t}^v(\bw_t^{(1)}) \Big)^2\nonumber\\
 &=\sum_{t=1}^k\|\bw_t^{(1)}\|_2^2\Big(\sum_{j=1}^m\epsilon_j
 \sum_{v\in N(j)}[g(\bL)]_{jv}h_{t}^v\big(\bw_t^{(1)}/\|\bw_t^{(1)}\|_2\big) \Big)^2.
\end{align*}
 By definition of the Frobenius norm, we conclude that the supremum of this over all $\bw_1^{(1)},\bw_2^{(1)},...,\bw_k^{(1)}$ such that  $\|\bW\|_{F}^2=\sum_{t=1}^k\|\bw_t\|_2^2\leq R^2$ must be attained when $\|\bw_{t_0}\|_2=R$ for some $t_0\in[k]$, and $\|\bw_t\|_2=0$ for all $t\neq t_0$. Therefore
\begin{align}\label{absot}
\mathbb{E}_{\epsilon}\mathop{\rm{sup}}_{\|\bW^{(1)}\|_{F}\leq R}
\Big\|\sum_{j=1}^m\epsilon_j   \sum_{v\in N(j)}[g(\bL)]_{jv} \bh_v \Big\|_2=
\mathbb{E}_{\epsilon}\mathop{\rm{sup}}_{\|\bw\|_2= R}
\sum_{j=1}^m\epsilon_j\Big(
\sum_{v\in N(j)}[g(\bL)]_{jv}h^v(\bw)\Big).
\end{align}
We only focus on the homogeneous setting for graph degree, namely, a common constant $q:=|N(j)|$  holds for all $j\in [m]$. Denote by $n_l(j)$  the  $l$-th  neighbor number of node $j$ by a given serial order, for all $l\in[q]$ and $j\in [m]$. With these notation, we can rewrite
\begin{align*}
\sum_{j=1}^m\epsilon_j\Big(
\sum_{v\in N(j)}[g(\bL)]_{jv}h^v(\bw)\Big)&=\sum_{l=1}^q\sum_{j=1}^m\epsilon_j\big(
[g(\bL)]_{jn_l(j)}h^{n_l(j)}(\bw)\big),\quad \forall\,\bw.
\end{align*}
Note that $\sup(\sum_la_l)\leq \sum_{l}\sup(a_l)$ for any $a_l\in \mathbb{R}$, this together with the above equality implies that 
\begin{align}\label{contra}
\mathbb{E}_{\epsilon}\mathop{\rm{sup}}_{\|\bw\|_2= R}
\sum_{j=1}^m\epsilon_j\Big(
\sum_{v\in N(j)}[g(\bL)]_{jv}h^v(\bw)\Big)\leq \sum_{l=1}^q
\mathbb{E}_{\epsilon}\mathop{\rm{sup}}_{\|\bw\|_2= R}
\sum_{j=1}^m\epsilon_j\big(
[g(\bL)]_{jn_l(j)}h^{n_l(j)}(\bw)\big).
\end{align}
For any fixed $l\in[q]$, we further  obtain that 
\begin{align*}
\mathbb{E}_{\epsilon}\mathop{\rm{sup}}_{\|\bw\|_2= R}
\sum_{j=1}^m\epsilon_j\big(
[g(\bL)]_{jn_l(j)}h^{n_l(j)}(\bw)\big)\leq 2\max_{j\in[m]}\big|[g(\bL)]_{jn_l(j)}\big|\times \mathbb{E}_{\epsilon}\mathop{\rm{sup}}_{\|\bw\|_2= R}
\sum_{j=1}^m\epsilon_j\big(h^{n_l(j)}(\bw)\big),
\end{align*}
where we use the contraction property of Rademacher average again. So far, combining  \eqref{radsecond}, \eqref{absot} and \eqref{contra}, we get
\begin{align}\label{comthree}
\widehat{\mathcal{R}}(\mathcal{F}_{D,R})
\leq \frac{4LD}{m}\sum_{l=1}^q \Big( \max_{j\in[m]}\big|[g(\bL)]_{jn_l(j)}\big|\times \mathbb{E}_{\epsilon}\mathop{\rm{sup}}_{\|\bw\|_2= R}
\sum_{j=1}^m\epsilon_j\big(h^{n_l(j)}(\bw)\big)\Big).
\end{align}
On the other hand, as the similar arguments in  \eqref{radfirst} and \eqref{supbound}, we have
\begin{align*}\label{inverse}
\mathbb{E}_{\epsilon}\mathop{\rm{sup}}_{\|\bw\|_2= R}
\sum_{j=1}^m\epsilon_j\big(h^{n_l(j)}(\bw)\big)&\leq 2L\mathbb{E}_{\epsilon}\mathop{\rm{sup}}_{\|\bw\|_2= R}
\sum_{j=1}^m\epsilon_j\Big(\sum_{r\in N(n_l(j))}[g(\bL)]_{n_l(j)r} \langle\bx_r, \bw\rangle\Big)\nonumber\\
&=2L\mathbb{E}_{\epsilon}\mathop{\rm{sup}}_{\|\bw\|_2= R}
 \Big\langle \sum_{j=1}^m\epsilon_j\sum_{r\in N(n_l(j))}[g(\bL)]_{n_l(j)r}\bx_r, \bw \Big\rangle,
\end{align*}
By Cauchy-Schwartz inequality and Jensen's inequality, this
 further is upper bounded by 
\begin{align}
  2LR \cdot \Big(\mathbb{E}_{\epsilon}\Big\|\sum_{j=1}^m\epsilon_j\sum_{r\in N(n_l(j))}[g(\bL)]_{n_l(j)r}\bx_r \Big\|^2_2\Big)^{\frac{1}{2}}
 =2LR \cdot\Big(\sum_{j=1}^m\Big\|\sum_{r\in N(n_l(j))}[g(\bL)]_{n_l(j)r}\bx_r \Big\|^2_2\Big)^{\frac{1}{2}},
\end{align}
where the last equality follows from  i.i.d. condition of Rademacher sequences with zero-mean. 
Next, we  denote by $g_{v}(\bL)\in \mathbb{R}^{q\times q}$ the sub-matrix of $g(\bL)$ whose row and column indices belong to the set  $\{r \in N(v)\}$. Let $\widetilde{\bX}_v=(\tilde{\bx}_1^T,...,\tilde{\bx}_q^T)^T \in \mathbb{R}^{q\times d}$ be the feature matrix of the nodes in $\mathcal{G}_v$. Then we notice
\begin{align*}
\Big\|\sum_{r\in N(n_l(j))}[g(\bL)]_{n_l(j)r}\bx_r \Big\|_2\leq \big\|\widetilde{\bX}_{n_l(j)}\big\|_2\max_{t\in[q]}\big\|[g_{n_l(j)}(\bL)]_{\cdot t}\big\|_2\leq \big\|\widetilde{\bX}_{n_l(j)}\big\|_2\big\|g_{n_l(j)}(\bL)\big\|_2,
\end{align*}
where the last inequality follows from the definition of $\|\bW \|_2$ for matrix. 
Denote by $\lambda_{\hbox{max}}(\mathcal{G})$  the maximum eigenvalue of $\bW$ of $g(\bL)$ over graph $\mathcal{G}$.
Then we have
\begin{align}\label{norm}
\Big\|\sum_{r\in N(n_l(j))}[g(\bL)]_{n_l(j)r}\bx_r \Big\|_2\leq\big\|\widetilde{\bX}_{n_l(j)}\big\|_2\big|\lambda_{\hbox{max}}(\mathcal{G}_{n_l(j)})\big| \leq\big\|\widetilde{\bX}_{n_l(j)}\big\|_2\big|\lambda_{\hbox{max}}(\mathcal{G})\big|
\end{align}
where we use the conclusion from \citet*{Laffey2008}:
$\big\|g_{v}(\bL)\big\|_2=|\lambda_{\hbox{max}}(\mathcal{G}_{v})|\leq |\lambda_{\hbox{max}}(\mathcal{G})|$ for all $v\in \mathcal{V}$. 
In addition, 
  the operator norm of matrix can be upper bounded by  
\begin{align*}
\big\|\widetilde{\bX}_{n_l(j)}\big\|_2=\sup_{\|\bw\|_2=1}\big\|\widetilde{\bX}_{n_l(j)}\bw\big\|_2\leq \big(\sum_{l=1}^q\|\tilde{\bx}_l\|^2_2\big)^{1/2}\leq B\sqrt{q},
\end{align*}
where $\|\bx_i\|_2\leq B$ by assumption. Then, we conclude from \eqref{norm} that
\begin{align*}
\Big\|\sum_{r\in N(n_l(j))}[g(\bL)]_{n_l(j)r}\bx_r \Big\|_2\leq B\sqrt{q}|\lambda_{\hbox{max}}(\mathcal{G})|,
\end{align*}
which in turn together with \eqref{inverse} implies, for all $l$
\begin{align}\label{partone}
\mathbb{E}_{\epsilon}\mathop{\rm{sup}}_{\|\bw\|_2= R}
\sum_{j=1}^m\epsilon_j\big(h^{n_l(j)}(\bw)\big)\leq 2LRB\sqrt{mq}|\lambda_{\hbox{max}}(\mathcal{G})|.
\end{align}	
Finally, combining	\eqref{comthree} and \eqref{partone}, we conclude that 
\begin{align}\label{lastone}
\widehat{\mathcal{R}}(\mathcal{F}_{D,R})
\leq 8L^2BDR|\lambda_{\hbox{max}}(\mathcal{G})|\sqrt{\frac{q}{m}}\sum_{l=1}^q  \max_{j\in[m]}\big|[g(\bL)]_{jn_l(j)}\big|.
\end{align}	
This completes the proof of Theorem 1. 	
\end{proof}

{\bf Proof for lower bound of Rademacher complexity}

\begin{proof}
By definition of the Rademacher complexity, it is enough to lower bound the complexity of some subset of $\mathcal{F}_{D,R}$, denoted by $\mathcal{F}'$. In particular, we focus on the class $\mathcal{F}_{D,R}'$ of graph neural networks over $\mathbb{R}$ of the form
\begin{align*}
\mathcal{F}_{D,R}':=\Big\{f(\bx_i)&=\sigma\Big(\sum_{t=1}^kw_t^{(2)}\sum_{v=1}^n[g(\bL)]_{iv}\times \sigma\big(\sum_{j\in N(v)}[g(\bL)]_{vj} 
\big\langle\bx_j, \bw_t^{(1)} \big\rangle \big) \Big),\,i\in[m],\nonumber\\
&\hspace*{0.7cm} \bW^{(1)}=(\bw^{(1)},{\bf 0},...,{\bf 0}),\;\bw^{(2)}=(w^{(2)},{\bf 0});\;\|\bw^{(1)}\|_{2}\leq R,\,|w^{(2)}|\leq D \Big\},
\end{align*}
where we choose  $\bW^{(1)}=(\bw^{(1)},{\bf 0},...,{\bf 0})$ so that only  the first column vector could be nonzero, in this case it holds $\|\bW^{(1)}\|_F=\|\bw^{(1)}\|_{2}\leq R$. Similarly, we only allow $\bw^{(2)}$ to vary in the first coordinate for simplifying proof. Furthermore, we take the linear activation $\sigma(s)=Ls$ as our choice.
In this setup, it holds that
\begin{align}\label{lowerb}
\widehat{\mathcal{R}}(\mathcal{F}_{D,R}')&\geq L^2\mathbb{E}_{\bepsilon}\sup_{\|\bw^{(1)}\|_2\leq R,\,|w^{(2)}|\leq D}\frac{1}{m}\Big|
\sum_{i=1}^m\epsilon_i\Big( w^{(2)}\sum_{v=1}^n[g(\bL)]_{iv}\times \big(\sum_{j\in N(v)}[g(\bL)]_{vj} \langle
\bx_{j},\bw^{(1)}\rangle\big)\Big)\Big|\nonumber\\
&= \frac{L^2D}{m} \mathbb{E}_{\bepsilon}\sup_{\|\bw^{(1)}\|_2\leq R}\Big|\Big\langle
\sum_{i=1}^m\epsilon_i\Big( \sum_{v=1}^n[g(\bL)]_{iv}\times \big(\sum_{j\in N(v)}[g(\bL)]_{vj} 
\bx_{j} \big)\Big),\,\bw^{(1)}\Big\rangle\Big|\nonumber\\
&=\frac{L^2RD}{m} \mathbb{E}_{\bepsilon}\Big\|
\sum_{i=1}^m\epsilon_i\Big( \sum_{v\in N(i)}[g(\bL)]_{iv}\times \big(\sum_{j\in N(v)}[g(\bL)]_{vj} 
\bx_{j} \big)\Big)\Big\|_2,
\end{align}
where the last step follows from the neighbor representation of graph shift operators, as well as the equivalent form of the
$L_2$-norm, that is, $\|\bs\|_2=\sup_{\|\bw\|_2=1}\langle  \bs, \bw\rangle$. Let $\be_1=(1,0,...,0)$ denote the standard unit vector in $\mathbb{R}^d$, and we assume that all the input data have the specific form $\bx_{j}=B\be_1$ for all $j\in [n]$. 
Then
\begin{align}\label{remx}
\Big\|
\sum_{i=1}^m\epsilon_i\Big( \sum_{v\in N(i)}[g(\bL)]_{iv}\big(\sum_{j\in N(v)}[g(\bL)]_{vj} 
\bx_{j} \big)\Big)\Big\|_2=B\Big|
\sum_{i=1}^m\epsilon_i\Big( \sum_{v\in N(i)}[g(\bL)]_{iv} \big(\sum_{j\in N(v)}[g(\bL)]_{vj} 
 \big)\Big)\Big|.
\end{align}
Note that, the exchange of summation  leads to the following equality
$$
\sum_{i=1}^m\epsilon_i\Big( \sum_{v\in N(i)}[g(\bL)]_{iv}\times \big(\sum_{j\in N(v)}[g(\bL)]_{vj} 
 \big)\Big)=\sum_{k=1}^q\sum_{t=1}^q[g(\bL)]_{kt}\Big(\sum_{i=1}^m\epsilon_i[g(\bL)]_{ik}\Big).  
$$
Suppose that the term $\sum_{t=1}^q[g(\bL)]_{kt}$ is invariant with $k$, denoted by $h_q(\bL)$.  Then
$$
\sum_{i=1}^m\epsilon_i\Big( \sum_{v\in N(i)}[g(\bL)]_{iv}\times \big(\sum_{j\in N(v)}[g(\bL)]_{vj} 
\big)\Big)=h_q(\bL)\sum_{i=1}^m\epsilon_i\Big(\sum_{k=1}^q[g(\bL)]_{ik}\Big).  
$$
Hence, this together with \eqref{remx} yields that 
\begin{align}\label{eachangeq}
\mathbb{E}_{\bepsilon}\Big\|
\sum_{i=1}^m\epsilon_i\Big( \sum_{v\in N(i)}[g(\bL)]_{iv} \big(\sum_{j\in N(v)}[g(\bL)]_{vj} \bx_{j}
\big)\Big)\Big\|_2=h^2_q(\bL)\mathbb{E}_{\bepsilon}\Big|\sum_{i=1}^m\epsilon_i\Big|=h^2_q\sqrt{m }.
\end{align}
Moreover, by  our choice of $\bx_{j}$ for all $j$ as above, we can check that
$$
|h_q(\bL)|=|\langle [g(\bL)]_{\cdot k}, {\bf 1}\rangle|=\big\|\widetilde{\bX}_q[g(\bL)]_{\cdot k} \big\|_2.
$$
As a consequence, combining \eqref{lowerb}, \eqref{remx} and \eqref{eachangeq}, we obtain
\begin{align*}
\widehat{\mathcal{R}}(\mathcal{F}_{D,R}')\geq
\frac{L^2BRD}{\sqrt{m}}\min_{k\in [q]}\Big\{ \big\|\widetilde{\bX}_q[g(\bL)]_{\cdot k} \big\|_2\sum_{t=1}^q[g(\bL)]_{kt}\Big\}.
\end{align*}
This completes the proof of Theorem \ref{lower}.  
\end{proof}

\section{Specific Examples}
In this section, we discuss the implications of our consistent results  of GCNs with one-hidden layer, specially in terms of graph models and graph convolution filters. 

 For any grpah model and graph convolution filer,  the dominant convergence rate in our result is of the order $O(\lambda_{\hbox{max}}(\mathcal{G})q^{3/2}m^{-1/2})$ up to all entries of $g(\bL)$. Note that the eigenvalue 
$\lambda_{\hbox{max}}(\mathcal{G})$ is determined by the specific filter, and the degree $q$ is determined by the graph model.

{\bf Unnormalized Graph Filters:}
One of the most popular graph filers is $g(\bL)=\bA+{\bf I}$, which has been used for GCNs in \cite{Kipf2016}. In this case,
Obviously, the eigen-spectrum of $g(\bL)$ and $q$ are completely determined by the graph structures. For Erdos-Renyi graphs with
edge probability $\Omega(\log (n)/n)$, the previous results in \cite{Krivelevich2003} showed that  $q=O(\log (n))$ and $|\lambda_{\hbox{max}}(\mathcal{G})|=O(\log (n))$ using the unnormalized filter. As a result, the generalization gap of such as GCN is of the order 
$O(\log^{5/2}(n)m^{-1/2})$ that has a weak dependence on $n$. In addition, for a regular graph with $q=O(1)$, we conclude  that
$|\lambda_{\hbox{max}}(\mathcal{G})|=O(1)$ in \cite{Bollobas1998}. This leads to  a graph size-independent generalization bound $O(m^{-1/2})$, which is in accordance with  same as typical Euclidean-based models.

{\bf Normalized Graph Filters:} Another graph filter  used widely in the graph network class is based on random walks: 
$g(\bL)=\bD^{-1}\bA+{\bf I}$ \citep*{Gilles2017}. Using such a filter, the eigenvalues of $\bD^{-1}\bA$ are bounded in the interval 
$[0,2]$, which implies $|\lambda_{\hbox{max}}(\mathcal{G})|=O(1)$ for any graph. Similar to the above, the model with the  Erdos-Renyi graph leads to  $q=O(\log (n))$, which further derives a generalization bound  $O(\log (n)^{3/2}m^{-1/2})$. Meanwhile, the regular model with $q=O(1)$ also leads to a generalization bound $O(m^{-1/2})$. Compared to the unnormalized case, the generalization bound of the normalized filter with Erdos-Renyi models is improved by $\log(n)$.

\section{Conclusions and Future Work}
We derived a sharp upper bound on the empirical Rademacher complexity of GCNs with one-hidden layer under the norm constraints of learned parameters. We also proved a lower bound that matches the upper one, and thereby shows the optimality of our result. 
Using the Rademacher complexity bound, we derived a generalization bound in terms of the degree  distribution of the graph and the spectrum of its graph convolution filter.

One future direction is to extend the  analysis to multi-hidden-layer graph networks, which  have been developed algorithmically for many practical problems. Another interesting direction is the related study on heterogeneous graphs under the neural network framework.

\bigskip



\end{document}